\documentclass[journal,hideappendix]{vgtc}        % final (journal style) without appendices
\onlineid{5636}

\vgtccategory{Research}

\title{GPUMAP: GPU-Accelerated Progressive Uniform Manifold Approximation and Projection}

\author{%
  Myeongwon Jung and
  Jaemin Jo
}

\authorfooter{
  \item Myeongwon Jung and Jaemin Jo are with Sungkyunkwan University.
        E-mail: \{mw.jung, jmjo\}@skku.edu. Jaemin Jo is the corresponding author.
}

\abstract{%
Dimensionality reduction (DR) techniques are fundamental to the analysis of high-dimensional data, yet their substantial computational cost limits their use in interactive settings.
To address this limitation, we present GPU-accelerated Progressive Uniform Manifold Approximation and Projection (\ours), an efficient progressive DR algorithm designed for interactive use.
We identify three atomic operations within UMAP and restructure them into a progressive pipeline. As part of this adaptation, we introduce modifications to UMAP, including operation queues, instance-wise local join, and scheduling strategies.
We further accelerate computation by leveraging GPU parallelism, enabling interactive analysis of million-scale datasets on commodity GPU hardware.
Finally, we employ our implementation to systematically study the behavior of progressive DR, investigating how projections evolve over time relative to the ground-truth projection and how biased data streams influence the quality and stability of the resulting projections.
}

\teaser{
  \centering
  \includegraphics[width=\linewidth, alt={A view of a city with buildings peeking out of the clouds.}]{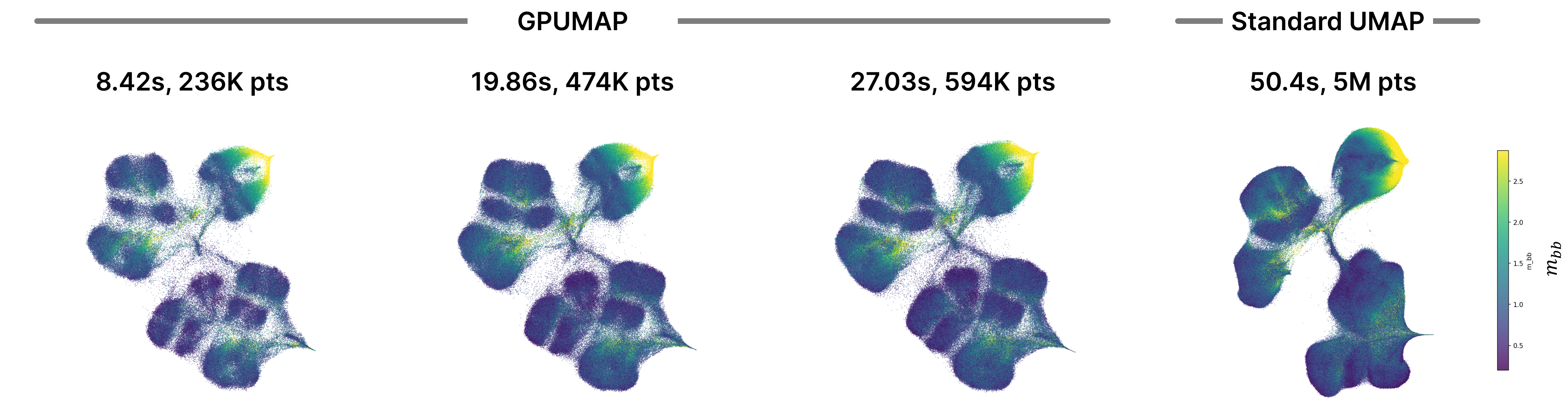}
  \caption{Intermediate projections produced by GPUMAP on the HIGGS dataset (5M points) before standard UMAP completes. GPUMAP produces projections at 8.42, 19.86, and 27.03 seconds as more points are inserted, while standard UMAP returns its complete projection at 50.4 seconds. Points are colored by the $m_\textit{bb}$ feature.}
  \label{fig:higgs}
}
\graphicspath{{figs/}{figures/}{pictures/}{images/}{./}} % where to search for the images

\usepackage{booktabs}                  % only used for the table example
\usepackage{lipsum}                    % used to generate placeholder text
\usepackage{mwe}                       % used to generate placeholder figures
\usepackage{ccicons}                   % package to be able to use icons from creative commons

\usepackage{mathptmx}                  % use matching math font

\usepackage{kotex}
\usepackage{xspace}
\usepackage{amsmath}
\newcommand{\ours}{GPUMAP\xspace}
\newcommand{\knn}{$k$NN\xspace}
\newcommand{\nnd}{NN-descent\xspace}
\usepackage{algorithm}
\usepackage{algpseudocode}
\usepackage{xcolor}
\usepackage[normalem]{ulem}
\usepackage{amssymb}

\usepackage{booktabs}   
\usepackage{siunitx}    
\usepackage{multirow}
\usepackage{mathrsfs}

\newcommand{\edit}[1]{{\color{black}{#1}}} % [리비전 - 본문에 리뷰 반영한 부분 파란색 표기]

\newif\ifshowdel
\showdelfalse
\newcommand{\del}[1]{\ifshowdel \textcolor{red}{#1} \else \ignorespaces \fi}

\newcommand{\rapids}{RAPIDS UMAP\xspace}
\newcommand{\ustd}{$\text{U}_\text{std}$\xspace}

\newcommand{\callIns}[1]{$\Call{Insert}{#1}$\xspace}
\newcommand{\callUpd}[1]{$\Call{Update}{#1}$\xspace}
\newcommand{\callEmb}[1]{$\Call{Embed}{#1}$\xspace}

\newcommand{\nupd}{$n_\text{upd}$\xspace}
\newcommand{\nemb}{$n_\text{emb}$\xspace}
\newcommand{\Qins}{$Q_\text{ins}$\xspace}
\newcommand{\Qupd}{$Q_\text{upd}$\xspace}
\newcommand{\Qemb}{$Q_\text{emb}$\xspace}
\newcommand{\Qembp}{$Q_\text{emb}^{\text{H}}$\xspace}
\newcommand{\tbudget}{$t_{\text{budget}}$\xspace}

\newcommand{\seq}{Sequential\xspace}

\newcommand{\insFoc}{Insert-Focused\xspace}
\newcommand{\updFoc}{Update-Focused\xspace}
\newcommand{\embFoc}{Embed-Focused\xspace}

\begin{document}

%%%%%%%%%%%%%%%%%%%%%%%%%%%%%%%%%%%%%%%%%%%%%%%%%%%%%%%%%%%%%%%%
%%%%%%%%%%%%%%%%%%%%%% START OF THE PAPER %%%%%%%%%%%%%%%%%%%%%%
%%%%%%%%%%%%%%%%%%%%%%%%%%%%%%%%%%%%%%%%%%%%%%%%%%%%%%%%%%%%%%%%

\firstsection{Introduction}

\maketitle

Dimensionality reduction (DR) techniques such as Uniform Manifold Approximation and Projection (UMAP)~\cite{mcinnes2018umap-software} have become a standard tool for exploring large-scale high-dimensional data.
\edit{
However, DR on large datasets is computationally expensive, dominated by neighborhood graph construction and embedding optimization.
Conventional implementations produce no projection until this computation completes: analysts cannot inspect the data, identify emerging structures, or judge whether a projection is useful for their analysis until the very end.
GPU-accelerated implementations~\cite{nolet2021bringing} shorten this wait but retain the same sequential execution model, and as datasets continue to grow, this all-or-nothing workflow makes DR increasingly difficult to use within an interactive exploration loop.
}

\edit{
Progressive Visual Analytics (PVA) offers a principled way to address this limitation by producing partial results while an analysis is still running~\cite{stolper2014progressive, fekete2016progressive}.
Rather than separating computation and inspection into distinct phases, PVA interleaves them, allowing users to examine emerging results before the underlying computation has finished.
Sampling also lowers the latency of obtaining a result, but it requires committing to a sample size in advance, before the analyst knows how much data the question at hand demands, and a sample that turns out to be too small cannot generally be extended without recomputing the result.
Applied to DR, this paradigm requires the system to produce a sequence of intermediate projections that are progressively refined as computation proceeds.
Progressive DR thus imposes two requirements: the first inspectable projection must appear with low latency, and the projection must be continuously updated throughout execution.

Several progressive DR methods have been proposed~\cite{jo2018panene, pezzotti2016approximated, ko2020progressive}, demonstrating that intermediate projections can be produced before the full computation completes.
However, their applicability to modern large-scale datasets remains limited.
These methods have primarily been evaluated on datasets containing tens of thousands of instances, leaving their scalability to million-scale data unexplored.
Furthermore, while these methods produce intermediate projections, the behaviors arising from such progressive execution have not been systematically studied.
For example, it remains unclear how rapidly the quality of intermediate projections converges to that of a sequentially computed projection and how distributing the computation budget among different operation types affects the projection quality.
}

\edit{
In this work, we present GPU-accelerated Progressive Uniform Manifold Approximation and Projection (\ours), an efficient and progressive implementation of the widely used DR method, UMAP~\cite{mcinnes2018umap-software}.
Existing approaches address only one side of the scalability problem: \rapids exploits GPU parallelism but preserves the sequential execution model, while existing progressive DR methods suffer from scalability limitations.
GPUMAP combines both directions, enabling progressive projection for million-scale data. 
% It decomposes the conventional UMAP pipeline into incremental operations for data insertion, neighborhood refinement, and embedding optimization, and coordinates them under a user-specified time budget. 
}

Our contributions are three-fold:
\begin{enumerate}
    \item We present \ours, an efficient progressive DR method that enables interactive analysis of large-scale high-dimensional data.
    \item We investigate how \ours approximates sequential DR projections in terms of runtime, nearest neighbor recall, and projection fidelity.
    \item We systematically evaluate progressive DR under diverse usage scenarios, including scheduling strategies, time budgets, and biased data streams.
\end{enumerate}

\section{Related Work}
\subsection{Uniform Manifold Approximation and Projection}
We build our method on UMAP, inspired by both the original Python implementation~\cite{mcinnes2018umap-software} and its GPU-accelerated RAPIDS implementation~\cite{nolet2021bringing}.
To clarify how we improve UMAP to support progressive computation, we briefly review its computational pipeline.
UMAP consists of two computation stages: $k$-Nearest Neighbor (\knn) graph construction and embedding optimization.

\textbf{\knn Construction.}
UMAP first constructs a \knn graph to capture topological relationships among data points in the input space.
Since exact \knn graph construction is expensive for large-scale datasets, UMAP mostly relies on approximate nearest neighbor algorithms such as Nearest-Neighbor Descent (\nnd)~\cite{dong2011efficient}. %, an approximate \knn algorithm that improves efficiency with only a minor loss in accuracy.
The intuition behind \nnd is that a neighbor of a neighbor is also likely to be a neighbor.
In \nnd, each point $v$ maintains a neighbor list $N_k(v)$, and forms a candidate set $C(v)=N_k(v)\cup N_k^{-1}(v)$, where $N_k(v)$ denotes the forward neighbors (points in $v$'s neighbor list) and $N_k^{-1}(v)=\{u \mid v \in N_k(u)\}$ denotes the backward neighbors (points that include $v$ in their neighbor lists).
\nnd then evaluates pairwise distances for all pairs $(u, w)$ such that $u,w\in C(v), u \neq w$, updating neighbor lists accordingly, which is called \textit{local join}.

\textbf{Embedding Optimization.}
Using the constructed \knn graph, UMAP computes a fuzzy topological representation by assigning a membership strength to each edge between a point $i$ and its neighbor $j$: $v_{j|i} = \exp({-\max(0, d(x_{i},x_{j})-\rho_{i})} / \sigma_{i})$,
where $x_i \in \mathbb{R}^D$ denotes the high-dimensional input coordinates of point $i$,
\del{$\rho_i$ denotes the distance from $x_i$ to its nearest neighbor (when a hyperparameter, local connectivity, is 1.0, which is the default), and $\sigma_i$ is a point-specific normalization factor.}
\edit{$\rho_i$ controls the local connectivity around $x_i$ and $\sigma_i$ is a point-specific normalization factor. With the default local connectivity of 1.0, $\rho_i$ corresponds to the distance from $x_i$ to its nearest neighbor.}
The final symmetric edge weight is given by $v_{ij} = (v_{j\mid i} + v_{i\mid j}) - v_{j\mid i} v_{i\mid j}$.

In the projection space, UMAP defines the low-dimensional edge weight as $w_{ij} = (1 + a\lVert y_{i} - y_{j} \rVert_{2}^{2b})^{-1},$
where $y_i \in \mathbb{R}^d$ ($d \ll D$) denotes the low-dimensional embedding coordinates of point $i$, $a$ and $b$ are hyperparameters that control the shape of the embedding distribution.
UMAP optimizes the embedding by minimizing the cross-entropy between $v_{ij}$ and $w_{ij}$:
\begin{equation} \label{eq:ce}
CE= \sum_{i \ne j} \left[ v_{ij} \cdot \log\left({v_{ij}}/{w_{ij}}\right) - (1 - v_{ij}) \cdot \log\left(({1 - v_{ij}})/({1 - w_{ij}})\right) \right].
\end{equation}
UMAP adopts a graph layout process with attractive and repulsive forces derived from the cross-entropy cost function.
\begin{equation}\label{eq:att}
    F_{att}(y_i,y_j,v_{ij}) = \frac{-2abd_{ij}^{2(b-1)}}{1 + d_{ij}^2} v_{ij} (y_i - y_j)
\end{equation}
\begin{equation}\label{eq:rep}
    F_{rep}(y_i,y_j,v_{ij}) = \frac{2b}{(\epsilon + d_{ij}^2)(1 + ad_{ij}^{2b})}(1 - v_{ij})(y_i - y_j)
\end{equation}   
where $d_{ij}=\|y_i - y_j\|_2$ and $\epsilon$ is a small constant to prevent division by zero.

From the implementation perspective, UMAP optimizes the embedding by computing attractive forces over \knn neighbors and repulsive forces over negatively sampled points, avoiding expensive all-pair computations.
The membership strength $v_{ij}$ is not directly multiplied into the gradient; instead, it is reflected as a sampling schedule:
\begin{equation} \label{eq:umap_schedule}
n_{ij}=n_{\text{epochs}}\frac{v_{ij}}{\max_{(p,q)}v_{pq}}, \ \text{epochs\_per\_sample}_{ij}=\frac{n_\text{epochs}}{n_{ij}},
\end{equation}
so that edges with larger weights are sampled more frequently.
When an edge $(i,j)$ is sampled, an attractive update is applied:
\begin{equation} \label{eq:attractive}
    \text{grad\_coeff}^+_{ij} = -2abd_{ij}^{2(b-1)}/({1 + ad_{ij}^{2b}}),
\end{equation}
and then applies multiple repulsive updates on randomly sampled negative points using
\begin{equation} \label{eq:repulsive}
    \text{grad\_coeff}^-_{ij} = 2\gamma b/((\varepsilon + d_{ij_k}^{2})(1 + ad_{ij_k}^{2b}))
\end{equation}
where $\gamma$ weights the repulsive term.
The coefficients are then applied to update the embedding coordinates as
\begin{equation} \label{eq:update}
    y_i \leftarrow y_i + \alpha \cdot \text{grad\_coeff}_{ij} \cdot (y_i - y_j),
\end{equation}
where $\alpha$ is the learning rate that linearly decays from an initial value $\alpha_0$ to zero over the course of optimization.

\subsection{Projection and Visualization for Large-scale Data}
To support the visual analysis of large-scale, high-dimensional data, prior work has explored methods to accelerate DR and reduce the computational burden.
Early efforts addressed the cost of \knn graph construction by employing approximate nearest neighbor (ANN) methods.
For example, A-tSNE~\cite{pezzotti2016approximated} computes an approximate \knn graph using a forest of randomized KD-Trees~\cite{silpa2008optimised}, while LargeVis~\cite{tang2016visualizing} constructs an approximate \knn graph using random projection trees~\cite{dasgupta2008random} followed by neighbor exploration to improve graph quality. 

In parallel, several studies aimed to reduce the embedding optimization process.
For example, Barnes-Hut-SNE~\cite{van2014accelerating} approximates long-range repulsive forces using a Barnes-Hut tree~\cite{barnes1986hierarchical}, reducing the complexity of optimization from quadratic to $O(N\log N)$, and Fit-SNE~\cite{linderman2019fast} further accelerates $t$-SNE by approximating the repulsive-force computation using interpolation and the Fast Fourier Transform (FFT). 
LargeVis also improves the embedding optimization process by adopting asynchronous stochastic gradient descent to minimize its objective in linear time.

More recently, these acceleration methods, including approximate \knn graph construction and scalable embedding optimization, have also been implemented on the GPU in open-source libraries such as RAPIDS cuML~\cite{raschka2020machine} and TorchDR~\cite{Van_Assel_TorchDR}.
These libraries provide GPU-accelerated implementations of several DR techniques, including UMAP.
However, such GPU-accelerated pipelines are typically benchmarked on high-end hardware such as an NVIDIA H100 GPU. On commodity GPUs, the achieved speedups are often insufficient to support interactive analysis of large-scale high-dimensional data.
\del{
In our work, we compare our implementation with a GPU-accelerated reference implementation, RAPIDS UMAP~\cite{raschka2020machine}, which we refer to as \textit{standard UMAP} hereafter.
}

Even with such acceleration, rendering and interactively exploring projections with millions of points remains computationally demanding.
Systems such as WizMap~\cite{wang2023wizmap} and Embedding Atlas~\cite{ren2025embedding} have therefore been proposed to support scalable rendering and interactive analysis of large projection spaces.
However, because these systems still depend on completed projections, users must wait before visual inspection becomes possible.

In this work, we extend UMAP with a GPU-accelerated progressive computation pipeline, enabling users to inspect intermediate projections that approximate the final result before optimization is fully complete. While our approach is compatible with scalable visualization systems such as Embedding Atlas, our primary focus is not on rendering techniques themselves, but on enabling progressive projection generation that supports early and continuous data analysis.

\edit{\subsection{Progressive Dimensionality Reduction}}
Progressive Visual Analytics (PVA) addresses the latency of long-running computations by exposing partial results early and continuously, allowing analysts to inspect evolving outputs and steer computation before final convergence~\cite{stolper2014progressive, fekete2016progressive}.
\del{Rather than blocking analysis until an exact result is available, PVA emphasizes meaningful intermediate results that are progressively refined over time, enabling interactive analysis even when data size or computational complexity makes conventional workflows too slow.}
\edit{Rather than blocking analysis until an exact result is available, PVA emphasizes meaningful intermediate results that are progressively refined over time.
Applied to DR, this principle requires producing inspectable projections before neighborhood construction and embedding optimization have fully converged.}

Prior work has explored progressive DR to reduce the initial latency.
DimXplorer~\cite{turkay2016designing} combines incremental PCA and mini-batch $k$-means to support progressively updated projections and clusters during exploratory analysis.
For non-linear DR, A-tSNE~\cite{pezzotti2016approximated} introduces controllable approximation and user-steerable local refinement to make $t$-SNE usable in a progressive setting.
PANENE~\cite{jo2018panene} further reduces startup cost through progressive approximate \knn computation and demonstrates Responsive $t$-SNE as one of its applications.
Progressive UMAP~\cite{ko2020progressive} extends this line of work to UMAP by incorporating progressive approximate \knn computation into the UMAP pipeline.

\edit{Neural parametric projection methods, such as Neural Network Projection (NNP)~\cite{espadoto2020deep}, parametric $t$-SNE~\cite{van2009learning}, and parametric UMAP~\cite{sainburg2021parametric}, also reduce projection latency by learning an explicit mapping that embeds new instances through a single forward pass. 
However, this speed applies only after training: they require a separate training stage on a representative subset of the data before any projection can be produced, so the latency of obtaining an initial result is shifted to, rather than eliminated by, this training stage.
They also rely on a fixed mapping that does not revise the positions of previously embedded points as new data arrive.
They therefore do not reduce the latency of the first projection nor progressively produce the projection of the current dataset, and are thus outside the scope of this work.}

\edit{
However, prior progressive DR methods have mainly been demonstrated at modest scales and with CPU-based implementations, leaving their practicality for large-scale interactive analysis unclear.
Moreover, it remains insufficiently understood how closely intermediate projections approximate the final projection obtained through sequential computation and how the quality of the progressive \knn graph and projection changes throughout progressive execution.
We address these gaps with a GPU-accelerated progressive UMAP method and a systematic analysis of its intermediate results.}

\vspace{3mm}
\section{GPUMAP}
% \todo{여기 섹션 위에 왜이렇게 세로 여백 공간이 좁을까요? 어딘가 vspace가 잘못 적용되고 있는거 아닌지?}

\begin{figure*}[t!]
    \centering
    \includegraphics[
        alt={},
        width=\linewidth
    ]{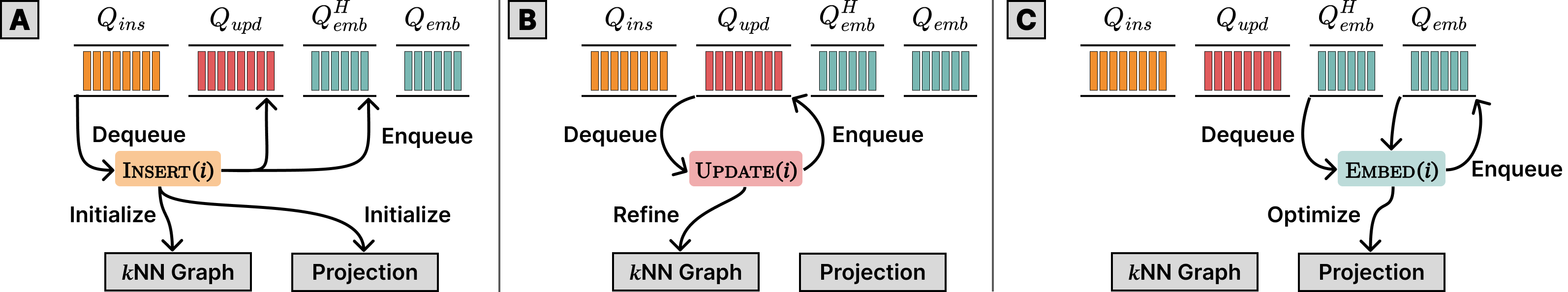}
    \caption{
    Overview of the three operations in the \ours pipeline.
    \textbf{(A)} \textbf{\textsc{Insert}}: points are dequeued from \Qins, inserted into the \knn graph with neighbors initialized via forward \nnd, and their initial embeddings are set based on their neighbors; the inserted points are then enqueued into \Qupd for neighbor refinement and \Qembp for initial embedding optimization with priority.
    \textbf{(B)} \textbf{\textsc{Update}}: points are dequeued from \Qupd to have their neighbors refined, and the \knn graph is refined via local join; points whose neighbors change are enqueued back into \Qupd for further refinement.
    \textbf{(C)} \textbf{\textsc{Embed}}: points are dequeued from \Qembp and \Qemb to optimize their embeddings; after optimization, points are enqueued into \Qemb for future revisits.    
    }
    \label{fig:Overview}
    % \vspace{-5mm}
\end{figure*}

\begin{algorithm}[t]
\caption{Progressive Execution in GPUMAP}
\label{alg:run}
\begin{algorithmic}[1]

\Function{Run}{$n_{\mathrm{ins}}, n_{\mathrm{upd}}, n_{\mathrm{emb}}$}
    \State \Call{InsertBatch}{$n_{\mathrm{ins}}$}
    \State \Call{UpdateBatch}{$n_{\mathrm{upd}}$}
    \State \Call{EmbedBatch}{$n_{\mathrm{emb}}$}
\EndFunction

\Statex

\Function{Run}{$t_{\mathrm{budget}}$}
    \State \parbox[t]{\dimexpr\linewidth-\algorithmicindent-1em\relax}{
    Allocate $t_{\mathrm{budget}}$ across operation types according to
    scheduling weights
    }
    \State Estimate the operation budget
           $(n_{\mathrm{ins}},n_{\mathrm{upd}},n_{\mathrm{emb}})$
    \State \Call{Run}{$n_{\mathrm{ins}},n_{\mathrm{upd}},n_{\mathrm{emb}}$}
\EndFunction

\end{algorithmic}
\end{algorithm}

\subsection{Overview of Progressive Execution}
To progressively compute UMAP, \ours decomposes the computation pipeline into a sequence of three types of operations that can be executed in an interleaved and controllable manner: \callIns{}, \callUpd{}, and \callEmb{} as illustrated in \cref{fig:Overview}.

\edit{An insertion operation, \callIns{i}, incorporates a new point $x_i$ into the current \knn graph, identifies approximate neighbors from the points inserted so far, and initializes its embedding. 
An update operation, \callUpd{i}, refines the local neighborhood structure around $x_i$ through a point-wise local join adapted from the \nnd~\cite{dong2011efficient}.
Finally, an embedding operation, \callEmb{i}, updates the embedding of $x_i$ using the latest graph state; the newly inserted points are processed with priority, while previously embedded points are revisited over time.
Together, these operations progressively expand data coverage, improve the \knn graph, and refine the projection.}

Each type of operation is managed by a dedicated queue on the host (i.e., CPU): the insertion queue (\Qins), update queue (\Qupd), and two embedding queues, \Qembp and \Qemb.
The priority embedding queue \Qembp contains the first embedding operation for each point and is processed with higher priority than \Qemb.

We present two execution interfaces to the client (\cref{alg:run}).
The first is $\textsc{Run}(n_{\text{ins}}, n_{\text{upd}}, n_{\text{emb}})$, where the client explicitly specifies an operation budget, requesting up to $n_{\text{ins}}$ insertion operations, $n_{\text{upd}}$ update operations, and $n_{\text{emb}}$ embedding operations to be processed from their respective queues.
\edit{Alternatively, the client can provide a wall-time budget \tbudget; GPUMAP allocates this budget across the three operation types according to a scheduling strategy and estimates the corresponding operation counts.
Although the operations are conceptually defined at the level of individual points, GPUMAP dequeues multiple operations of the same type and executes them as GPU batches for efficiency.
After each batch execution, GPUMAP produces the updated projection along with execution statistics, such as the actual time spent on each type of operation.
We describe the procedures and design rationale in the following sections, while providing complete pseudocode in the supplementary material.}
% We also provide an open-source implementation of \ours\footnote{The source code will be made publicly available upon acceptance.}.
% \todo{소스 코드 억셉되면 보여주겠다 이게 지금 구현이랑 두 번 나와요. 둘 중 하나를 지워서 한번만 얘기합시다.}

\subsection{The Insertion Operation}
\begin{algorithm}[t]
\caption{Insertion Operation}
\label{alg:insert}
\begin{algorithmic}[1]

\Function{Insert}{$i$}
    \State $\mathcal{N}_k(i) \gets \Call{SeedNeighbors}{i, \mathcal{N}}$
    \State $A \gets \Call{RefineForwardNeighbors}{i, \mathcal{N}}$
    \State $y_i \gets \Call{InitializeEmbedding}{\mathcal{N}_k(i), \mathbf{Y}}$
    \State $\Call{Enqueue}{Q_\text{upd}, \{i\} \cup N_k(i) \cup A}$
    \State $\Call{Enqueue}{Q_\text{emb}^{\text{H}}, i}$
\EndFunction

\Statex
\Function{RefineForwardNeighbors}{$i, \mathcal{N}$}
    \State $A \gets \emptyset$
    \For{$t \gets 1$ \textbf{to} $T$}
        \State $C \gets \bigcup_{j \in N_k(i)} N_k(j) \setminus \{i\}$
        \ForAll{$c \in C \setminus \{i\}$}
            \State $A \gets A \cup
                \Call{UpdateNeighbors}{i, c, \mathcal{N}}$
        \EndFor
    \EndFor
    \State \Return $A$
\EndFunction

\Statex

% \Function{InsertBatch}{$n_{\mathrm{ins}}$}
%     \State $B \gets$ dequeue up to $n_{\mathrm{ins}}$ points from \Qins
%     \State execute \Call{Insert}{$i$} for all $i \in B$ in parallel
% \EndFunction
\Function{InsertBatch}{$n_\mathrm{ins}$}
    \State $B_\mathrm{ins} \gets \Call{Dequeue}{Q_\mathrm{ins}, \min(n_\mathrm{ins}, |Q_\mathrm{ins}|)}$
    \ForAll{$i \in B_\mathrm{ins}$ \textbf{in parallel}}
        \State \Call{Insert}{$i$}
    \EndFor
\EndFunction

\end{algorithmic}
\end{algorithm}
An insertion operation, \callIns{i}, adds a data point $x_i$ to the \knn graph and constructs its initial neighbor list using the points inserted so far.
Given a dataset of $M$ points, \callIns{1}$, \ldots,$ \callIns{M} are initially enqueued in \Qins and await execution by the $\textsc{Run}$ call.
\edit{The goal of each insertion is not to fully refine the neighborhood structure around $x_i$, but to rapidly establish a plausible local neighborhood.
As illustrated in \cref{alg:insert}, the operation consists of three main steps: neighbor seeding, forward neighbor refinement, and embedding initialization, followed by scheduling the affected points for subsequent refinement.}

\textbf{Neighbor Seeding.}
To initialize the neighbor list of the newly inserted point $i$, \ours first randomly samples half of the candidates from the most recently inserted points and the remaining half from all points inserted so far.
The original \nnd algorithm randomly selects a seed set of neighbors from the entire dataset, but this is infeasible in progressive execution, as the algorithm has no access to points that have not yet been inserted.
A straightforward alternative is to sample uniformly from the set of already inserted points, but this introduces bias: earlier points are more likely to be selected as initial neighbors, causing the graph to grow around the initial structure.
To mitigate this bias, we maintain a \textbf{recent pool} ($\mathcal{W}$), defined as a sliding window over the most recent $W$ batches ($W=3$ by default).
For example, after inserting nine batches $B_1$ through $B_9$, when inserting the 10th batch $B_{10}$, the seed neighbors for each point in $B_{10}$ are sampled such that half are drawn uniformly from all inserted points ($B_1$ to $B_{10}$), and the other half are drawn from the recent pool ($B_8$ to $B_{10}$).

\textbf{Forward Neighbor Refinement.}
The random initial neighbor list of point $i$ should be refined. 
The original \nnd performs this refinement by repeatedly running local join; for example, it computes distances for every pair within the union of forward neighbors $N_k(i)$ and backward neighbors $N_k^{-1}(i)$.
We instead consider only second-order forward neighbors as candidates, i.e., $C = \bigcup_{j \in N_k(i)} N_k(j) \setminus \{i\}$, the neighbors of $i$'s current neighbors.
This is because (1) our goal is the rapid identification of plausible neighbors for $i$ not global graph refinement and (2) computing backward neighbors incurs larger computational overhead. 
The local join on $i$, including backward neighbors, is deferred to the subsequent \callUpd{} operation.
For each candidate in $C$, we compute the distance to $i$ and update the neighbor lists of both points accordingly.
Any point whose neighbor list changes is added to the affected set $A$, which records points that require subsequent \callUpd{} processing.
% \todo{여기서 A의 정의가 없어요}
This refinement is repeated $T=10$ times by default.

\textbf{Embedding Initialization.}
Once the neighbor list of point $i$ has been refined, we initialize its embedding as an exponentially weighted average of its neighbors' embeddings, where the weights are determined by their distances. This is motivated by UMAP’s use of an exponential kernel to model membership strength.

\textbf{Queueing.}
\edit{At the end of the insertion operation, we enqueue \callUpd{} operations for points affected by the insertion.
These points include the newly inserted point $i$, its neighbors in $N_k(i)$, and the existing points whose neighbor lists were updated to include $i$ during forward neighbor refinement (Line 5 in \cref{alg:insert}).
By restricting subsequent refinement to these points, GPUMAP avoids unnecessary computation on unaffected regions of the neighbor graph.
Finally, an \callEmb{i} operation is enqueued into the priority embedding queue \Qembp, so that the embedding of $i$ is optimized with higher priority, as it has only been initialized but not yet optimized.}

\subsection{The Update Operation}
\begin{algorithm}[t]
\caption{Update Operation}
\label{alg:update}
\begin{algorithmic}[1]
\Function{Update}{$i$}
    \State $C_{\mathrm{new}}, C_{\mathrm{old}}
        \gets \Call{SampleCandidates}{i, \mathcal{N}}$
    \State $A \gets
        \Call{LocalJoin}{C_{\mathrm{new}}, C_{\mathrm{old}}}$
    \State $\Call{Enqueue}{Q_\text{upd}, A}$
\EndFunction
\Statex

\Function{LocalJoin}{$C_{\mathrm{new}}, C_{\mathrm{old}}$}
    \State $A \gets \emptyset$
    \For{each $(c_1, c_2) \in \Call{Pairs}{C_{\mathrm{new}}, C_{\mathrm{old}}}$}
        \State $A \gets A \cup \Call{UpdateNeighbors}{c_1, c_2, \mathcal{N}}$
    \EndFor
    \State \Return $A$
\EndFunction

\Statex

\Function{UpdateBatch}{$n_\mathrm{upd}$}
    \State $n_\mathrm{left} \gets n_\mathrm{upd}$
    \While{$|Q_\mathrm{upd}| \geq \delta \cdot M'$ \textbf{and} $n_\mathrm{left} > 0$}
        \State $B_\mathrm{upd} \gets \Call{NextBatch}{Q_\mathrm{upd}, n_\mathrm{left}, \mathcal{M}'}$
        \ForAll{$i \in B_\mathrm{upd}$ \textbf{in parallel}}
            \State \Call{Update}{$i$}
        \EndFor
        \State $n_\mathrm{left} \gets n_\mathrm{left} - |B_\mathrm{upd}|$
    \EndWhile
    % \State \Return $n_\mathrm{upd} - n_\mathrm{left}$
\EndFunction

\end{algorithmic}
\end{algorithm}
\edit{An update operation, \callUpd{i}, refines the \knn graph around point $i$ by running the local join algorithm \cite{dong2011efficient}, propagating neighborhood information to points not directly touched by insertion.
Rather than sweeping over all points at once as in the original \nnd, we decompose this into independent \callUpd{i} calls, each queued in \Qupd and triggered only for points whose neighbor lists changed during a preceding \callIns{} or \callUpd{}.
\cref{alg:update} illustrates the procedure for a single point.}

\edit{\textbf{Candidate Sampling.}
For each \callUpd{i}, GPUMAP collects nearby points that may reveal new neighbor relationships through a local join. 
These candidates are drawn from the forward and sampled backward neighbors of $i$. 
Following the original \nnd, GPUMAP distinguishes candidates newly introduced to the local join from those already processed. 
This distinction allows the subsequent local join to focus on pairs involving new neighborhood information and avoid repeating comparisons between previously processed candidates. 
We denote the resulting sets by $C_{\mathrm{new}}$ and $C_{\mathrm{old}}$.}

\edit{\textbf{Local Join.}
\ours evaluates candidate pairs consisting of at least one new candidate: new--new and new--old pairs ($\Call{Pairs}{}$ in \cref{alg:update}); old--old pairs are skipped, since both points were already compared in a preceding update.
For each pair, \ours computes the distance between the two points and updates their neighbor lists if either point provides the other with a closer neighbor. 
% All points whose neighbor lists change are collected in a set $A$.
Any point whose neighbor list changes is likewise added to $A$.}
% \todo{여긴 A가 나오네요?}.

\edit{\textbf{Queueing.}
After each \callUpd{i}, GPUMAP enqueues the affected points in $A$ into \Qupd.
This allows newly discovered neighbor relationships to propagate incrementally through subsequent updates while avoiding computation on points whose neighbor lists remain unchanged.}

\edit{\textbf{Batch Scheduling.}
Beyond a single point, $\Call{UpdateBatch}{}$ schedules \callUpd{} calls across many points within a fixed budget \nupd per round.
Each round, it draws a batch of points via $\Call{NextBatch}{}$ and runs \callUpd{} on them in parallel.
$\Call{NextBatch}{}$ dequeues as many points as available from \Qupd.
If fewer points remain than the round's budget, it supplements the batch with points randomly sampled from $\mathcal{M}' \setminus B_\mathrm{upd}$, where $\mathcal{M}'$ denotes the set of all points inserted so far.
This covers points that still carry unprocessed \emph{new} flags but were missed by backward sampling.
The process repeats until either the budget \nupd is exhausted or $|Q_\text{upd}|$ falls below the convergence threshold $\delta \cdot M'$ (we use $\delta = 0.001$, following the original \nnd).}

\begin{algorithm}[t]
\caption{Embedding Operation}
\label{alg:emb}
\begin{algorithmic}[1]

\Function{Embed}{$i$}
    \State \Call{AddPerturbation}{$y_i$} \Comment{add small perturbation to $y_i$}
    \State $E_i \gets \{(i,j,v_{ij}) \mid j \in N_k(i), v_{ij} \in \mathcal{G}\}$
    \State \Return $E_i$
\EndFunction
\Statex
\Function{EmbedBatch}{$n_\mathrm{emb}$}
    \State $B_\mathrm{emb} \gets \Call{Dequeue}{Q_\text{emb}^{\text{H}},\ n_{\mathrm{emb}}}$
    \State $B_\mathrm{emb} \gets B_\mathrm{emb} \cup \Call{Dequeue}{Q_\text{emb},\ \max(0,\ n_{\mathrm{emb}}-|B_\mathrm{emb}|)}$
    \State $\mathcal{G} \gets \Call{ComputeFuzzyGraph}{\mathcal{N}}$
    \State $\mathcal{E} \gets \emptyset$
    \ForAll{$i \in B_\mathrm{emb}$ \textbf{in parallel}}
        \State $\mathcal{E} \gets \mathcal{E} \cup \Call{Embed}{i}$
    \EndFor
    \For{$e \gets 1$ to $e_{max}$} \Comment{synchronize epoch progression}
        \ForAll{$(i,j,v_{ij}) \in \mathcal{E}$ \textbf{in parallel}}
            \State $\Call{UpdateEdge}{i, j, v_{ij}, e}$
        \EndFor
    \EndFor
    \ForAll{$k \in \mathcal{M}' \setminus B_\mathrm{emb}$ \textbf{in parallel}}
        \State $\Call{PropagatePosition}{k, Y, \mathcal{G}}$
    \EndFor
    \State $\Call{Enqueue}{Q_\text{emb},B_\mathrm{emb}}$
\EndFunction
\end{algorithmic}
\end{algorithm}

\subsection{The Embedding Operation}
An embedding operation. \callEmb{i}, refines the low-dimensional position of the point $i$ based on the current \knn graph extended and refined by the \callIns{} and \callUpd{} operations.
Unlike standard UMAP, which optimizes the embedding over all points using the full dataset, our progressive setting allocates a fixed computational budget of \nemb points to each run.
Accordingly, each embedding run processes at most \nemb points, making the execution time controllable.
The overall procedure for executing embedding operations is summarized in \cref{alg:emb}.
% \todo{밑에 함수들을 설명하는데 막상 algorithm 4 인용이 어딘가에 없음}

An alternative design for allocating budgets would be to partition embedding updates at the epoch level, executing a few epochs for all points.
While we initially explored this approach, it yielded suboptimal results.
In particular, interrupting the optimization before completing all $e_\text{max}$ epochs led to unstructured and scattered embeddings.
Instead, GPUMAP completes all epochs for the selected points and propagates the resulting changes to the remaining points, producing a more coherent partial layout.

\edit{\textbf{Batch-wise Optimization.}
Although \callEmb{i} is defined for an individual point, embedding operations are executed jointly in GPU batches.
Given a budget of \nemb, \Call{EmbedBatch}{} selects up to \nemb points from the embedding queues and constructs the fuzzy graph $\mathcal{G}$ from the current \knn graph following standard UMAP.
For each selected point $i$, \callEmb{i} identifies its incident fuzzy graph edges $\mathcal{E}_i$ as the targets of direct optimization.
\Call{EmbedBatch}{} combines the edge sets from all selected points and optimizes their union over a single shared loop of $e_{\max}$ epochs.
Finally, \Call{PropagatePosition}{} updates the remaining points to reflect the layout changes produced by the batch.
This formulation allows GPUMAP to allocate its embedding budget by point while retaining the edge-parallel optimization of standard UMAP.}

\edit{\textbf{Maintaining Embedding Quality.}
We employ two mechanisms to maintain embedding quality across repeated, partial optimization runs.
First, repeatedly optimizing the same points can lead to poor local optima and excessive contraction of local structures.
To alleviate this issue, we add a small Gaussian perturbation to the points before embedding operations ($\Call{AddPerturbation}{}$ in \cref{alg:emb}).
Second, because points outside $B_{\mathrm{emb}}$ receive no direct edge updates, we update their positions by repositioning to the weighted average of their neighbors' embeddings ($\Call{PropagatePosition}{}$ in \cref{alg:emb}), which is a cheap operation.}

\textbf{Queueing.}
\ours maintains two embedding queues, \Qembp (high priority) and \Qemb (low priority), to control the order in which points undergo optimization.
Without this separation, newly inserted points would be placed at the end of a single queue and could be delayed for several runs, resulting in poor embedding quality.
Points are drawn from \Qembp first, and 
if the embedding budget $n_{\mathrm{emb}}$ permits, additional points are drawn from \Qemb.
After each embedding run, all processed points, including those from \Qembp, are inserted into \Qemb again to enable continued refinement if either \Qins or \Qupd is non-empty.
This design can be viewed as a round-robin update scheme over all inserted points, ensuring that each point is revisited over time.

\subsection{Execution Span Approximation}

While \ours supports incremental computation of UMAP by executing a specified number of operations of each type, users may instead wish to run the computation for a fixed time budget (e.g., 10 seconds), allowing the algorithm to determine how many operations can be executed within that time.
To translate a physical time budget into the number of operations, we adopt a simple linear regression strategy similar to that used in ProgressiVis~\cite{fekete2015progressivis}. 
Specifically, we record the execution time of each operation type over the last five runs and use these measurements to predict the per-operation cost in the current run, while reserving a small portion of the budget to account for estimation error (\cref{alg:run}).

When no execution history is available, runtime estimation must rely on an initial calibration.
A straightforward approach is to perform several warm-up runs with a small number of operations; however, this is inefficient because, for a fixed hardware configuration and dataset dimensionality, the runtime characteristics remain largely stable.
Therefore, we store runtime statistics in a local database and reuse them in future runs. 
If no such record exists (e.g., during the first run on a given dataset and hardware configuration), we perform a short calibration procedure that begins with a very small number of operations (e.g., 1,000 operations per type) and repeatedly doubles the number of operations to initialize the regression model.

\subsection{Scheduling}\label{sect:scheduling}
Given a fixed time budget $t_{\text{budget}}$, \ours determines how to allocate this budget across the three operation types, which we refer to as a \textbf{scheduling strategy}.
As a baseline, we define \textbf{\seq}, a non-progressive scheduling strategy equivalent to invoking $\textsc{Run}(\infty, \infty, \infty)$, thereby ignoring the budget.
Under \seq, all \callIns{} operations are completed first, followed by \callUpd{}, and finally \callEmb{}, making it the execution mode in \ours that most closely resembles standard UMAP.
We use this strategy both as a sanity check and as a controlled baseline for runtime comparison with progressive DR.

We parametrize each scheduling strategy by a weight vector 
$W = (w_\mathrm{ins}, w_\mathrm{upd}, w_\mathrm{emb})$, which determines how the budget is distributed across \callIns{}, \callUpd{}, and \callEmb{}.
We define a baseline scheduling strategy \textbf{\updFoc} with $W = (1, 2, 1)$; this update-biased weighting reflects our empirical observation that the quality of the \knn graph is the dominant factor influencing projection quality.
We further define two variants, \textbf{\insFoc} and \textbf{\embFoc}, which increase the relative emphasis on \callIns{} and \callEmb{}, respectively, by doubling their weights, resulting in $W = (2, 1, 1)$ and $W = (1, 1, 2)$.

\subsection{Implementation}
\ours is implemented in CUDA/C++ and exposed to Python via pybind11~\cite{pybind11}\footnote{The source code will be made publicly
available upon acceptance.}.
We build upon the NVIDIA RAPIDS cuML library~\cite{raschka2020machine}, leveraging its parallel primitives, including GPU kernels for \nnd and UMAP, while introducing our modifications to enable progressive execution.
At present, \ours supports single-GPU execution only and does not include multi-GPU support.

\section{Evaluation}

\edit{Although several progressive DR methods have been proposed~\cite{jo2018panene, pezzotti2016approximated, ko2020progressive}, it remains insufficiently understood how the quality of their intermediate results evolves over time and how progressive execution affects computational performance.
In this section, we address this gap by using \ours as a concrete realization of progressive DR and standard UMAP, denoted as \ustd, as its non-progressive counterpart.
We first verify that the progressive decomposition of UMAP preserves the final neighborhood and projection quality across a diverse collection of datasets.
We then examine how these qualities change throughout execution under different scheduling strategies and time budgets.
Next, we compare \ours with Progressive UMAP~\cite{ko2020progressive} to assess how efficiently each method improves projection fidelity and neighborhood quality over time.
Finally, we examine the robustness of progressive execution under biased data insertion orders.}

\begin{table}[t]
\centering
\renewcommand{\arraystretch}{1.1}
\caption{Summary of datasets used in the evaluation.\vspace{-2mm}}
% \scriptsize
\begin{tabular}{@{\hspace{3pt}}p{3.5cm} c @{\hspace{30pt}} c@{\hspace{3pt}}}
\toprule
\textbf{Dataset} & \textbf{Size} ($M \times$ dims) & \textbf{Data Type} \\
\midrule
% \multicolumn{3}{l}{\textit{Real-world datasets}} \\
EMNIST Digits~\cite{cohen2017emnist}       & \phantom{0,}280,000 $\times$ 784  & Image \\
Covertype~\cite{covertype_31}       & \phantom{0,}581,012 $\times$ 54  & Table \\
SVHN~\cite{netzer2011reading}       & \phantom{0,}630,420 $\times$ 768  & Image \\
SIFT1M~\cite{jegou2010product}       & 1,000,000 $\times$ 128  & Image \\
MARC~\cite{keung2020multilingual} & 1,200,000 $\times$ 768 & Text \\
Mouse Brain~\cite{gse93421_mouse_brain}     & 1,306,127 $\times$ 50\phantom{0}  & Genomic  \\
LAION~\cite{schuhmann2021laion}  & 2,000,896 $\times$ 512  & Image \\
Google News~\cite{mikolov2013distributed}       & 3,000,000 $\times$ 300  & Text \\
% HIGGS~\cite{higgs_280} & 5,000,000$\times$ 28\phantom{0} & Table \\
\midrule
% \multicolumn{3}{l}{\textit{Synthetic datasets}} \\% todo: 확정되면 추가
% Swiss roll       & 1,000,000 $\times$ 3\phantom{00}  & Synthetic \\
Spheres        &     1,000,000 $\times$ 3\phantom{00}  & Synthetic \\
Cylinder       &     1,000,000 $\times$ 3\phantom{00}  & Synthetic \\
\bottomrule
\end{tabular}
\label{tab:datasets}
\end{table}

\begin{table}[t]
\centering
\caption{
  Results of sanity checks.
  % We report the number of points, number of runs, total runtime, final \knn graph recall, fidelity, ground-truth fidelity, and normalized fidelity.
}
\label{tab:sanity_check}
\setlength{\tabcolsep}{5pt}
\begin{tabular}{lrrr}
\toprule
\multirow{2}{*}{\textbf{Dataset}}
  & \multirow{2}{*}{\textbf{Time (s)}}
  & \multirow{2}{*}{\textbf{Final Recall}}
  & \multirow{2}{*}{\textbf{Norm. Fidelity}} \\
\\
\midrule
EMNIST Digits & $19.97{\scriptstyle\pm0.52}$ & $0.99{\scriptstyle\pm0.00}$ & $0.98{\scriptstyle\pm0.02}$ \\
Covertype & $27.41{\scriptstyle\pm0.66}$ & $1.00{\scriptstyle\pm0.00}$ & $1.00{\scriptstyle\pm0.00}$ \\
SVHN & $47.01{\scriptstyle\pm0.65}$ & $0.97{\scriptstyle\pm0.00}$ & $0.94{\scriptstyle\pm0.02}$ \\
SIFT1M & $56.10{\scriptstyle\pm1.22}$ & $0.93{\scriptstyle\pm0.00}$ & $0.86{\scriptstyle\pm0.00}$ \\
MARC & $95.81{\scriptstyle\pm0.45}$ & $0.84{\scriptstyle\pm0.00}$ & $0.74{\scriptstyle\pm0.01}$ \\
Mouse Brain & $68.88{\scriptstyle\pm0.70}$ & $0.95{\scriptstyle\pm0.00}$ & $1.00{\scriptstyle\pm0.00}$ \\
LAION & $141.33{\scriptstyle\pm0.76}$ & $0.92{\scriptstyle\pm0.00}$ & $0.96{\scriptstyle\pm0.02}$ \\
Google News & $197.86{\scriptstyle\pm1.14}$ & $0.80{\scriptstyle\pm0.00}$ & $0.96{\scriptstyle\pm0.01}$ \\

\bottomrule
\end{tabular}
\end{table}

\subsection{Experimental Setup}
\textbf{Datasets.}
\edit{We evaluate \ours on a collection of real-world datasets spanning image, tabular, genomic, and text data, with dataset sizes ranging from hundreds of thousands to several million instances (\cref{tab:datasets}).
For the robustness assessment, we additionally use two synthetic datasets, Spheres and Cylinder, whose geometric structures and insertion orders can be explicitly controlled.
Detailed descriptions and preprocessing procedures for each dataset are provided in the supplementary material.}

\edit{\textbf{Configurations.}
We use two baseline implementations.
First, we use RAPIDS UMAP~\cite{nolet2021bringing} as the specific implementation of \ustd.
Second, we use Progressive UMAP~\cite{ko2020progressive} as the prior progressive baseline.
We do not use the official UMAP implementation~\cite{mcinnes2018umap-software}, as it is neither GPU-accelerated nor progressive.
For hyperparameters shared with \ustd, we use the default UMAP settings across experiments, except that cosine distance is used for the text embedding datasets.
All UMAP-based methods use random initialization rather than spectral initialization, so that the comparisons do not rely on a global initialization computed from the full dataset.
For \ours-specific parameters, we use a fixed configuration across datasets unless otherwise specified; the complete parameter settings are reported in the supplementary material.
For experiments that examine scheduling behavior, we use the four strategies defined in \cref{sect:scheduling}.}
All experiments were conducted on a desktop equipped with an AMD Ryzen 9 5950X CPU, 128GB of DDR4 RAM, and an NVIDIA RTX A6000 GPU with 48GB of VRAM.

\edit{\textbf{Time Budgets.}
For progressive execution, we primarily use a per-run time budget of $t_{\mathrm{budget}}=10$s.
We additionally evaluate a 5-second budget to examine how more frequent updates affect progressive behavior and computational throughput.}

\begin{figure*}[t!]
    \centering
    \includegraphics[
        alt={},
        width=\linewidth
    ]{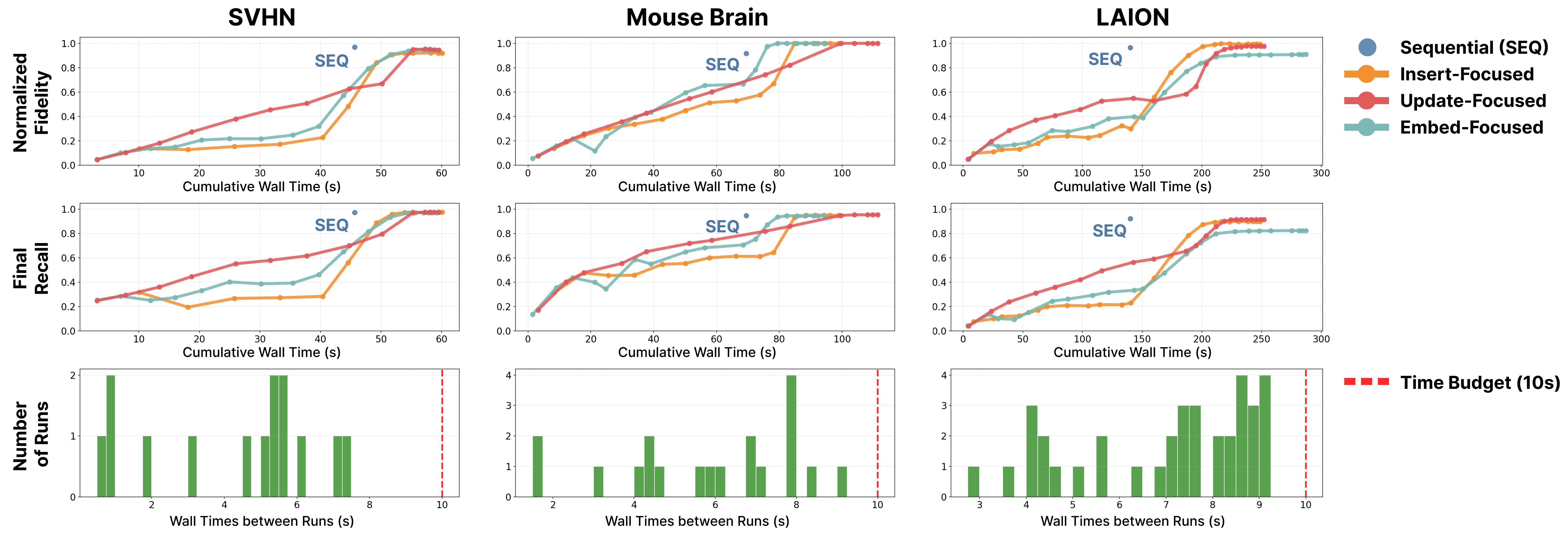}
    \caption{
Comparison of scheduling strategies in terms of normalized fidelity, final recall, and per-run wall-time distribution.
In the top two rows, blue points indicate the final result of the sequential baseline (\seq).
The bottom row shows the per-run wall-time distribution of the \updFoc strategies under a 10-second budget, with the red dashed line indicating the time budget.
% \todo{sequential이 그래프에서 점 하나로만 보여서 모르는 사람이 찾기가 어려울 수도 있을 것 같아요. 점 밑에 작게라도 SEQ 이런걸 적어주는게 어때요?}
    }
    \label{fig:strategies}
\end{figure*}

\subsection{Metrics}
We evaluate GPUMAP in terms of execution time, nearest neighbor recall, and projection fidelity.

\textbf{Execution Time.}
We measure the overall wall-clock time measured from the Python interface after all operations for each $\Call{Run}{}$ call have completed.
To isolate the performance of the algorithm itself, we assume a setting without a client; accordingly, external overheads such as network transmission and client-side rendering are excluded.

\textbf{Nearest Neighbor Recall.}
We evaluate the quality of the constructed \knn graph using nearest neighbor recall.
We measure \textbf{final recall}, which compares the current \knn graph against the ground-truth \knn graph over the full dataset.
Note that the final recall is strictly upper-bounded by the proportion of data inserted so far; since the \knn graph is built over only the points inserted up to the current iteration, it cannot capture neighbors that have yet to be inserted.

\textbf{Normalized Fidelity.}
We quantify how faithfully a progressive projection preserves the structure of the sequentially computed projection (hereafter referred to as the ground-truth projection), which we denote as the fidelity of the progressive projection.
We initially considered standard DR quality metrics such as Trustworthiness and Continuity~\cite{venna2006local}.
However, these metrics are not suitable for our setting for two reasons.
First, they are designed to measure the alignment between high- and low-dimensional spaces, rather than the similarity between two projections.
Second, similar scores under these metrics do not necessarily imply that the projections themselves are structurally similar.

We found that measuring the fidelity of a progressive projection is not yet well established, and identified three fundamental challenges.
First, at any intermediate iteration, only a subset of $M' \leq M$ data points have been projected, whereas the ground-truth projection covers all $M$ points. 
Comparing the two projections, therefore, requires a way to fill in the missing projections of $M - M'$ points.
Second, two projections of the same data may differ substantially in their absolute coordinates despite preserving similar structures, making direct coordinate- or pixel-level comparisons inappropriate.
Finally, UMAP and its progressive variants are stochastic algorithms due to the use of approximate \knn, stochastic gradient descent, and negative sampling.
Therefore, a single reference projection is an insufficient ground truth.

To address the first challenge, we employ density-based filling; we fill the unprojected points by sampling positions proportional to the empirical density of the projected points.
Concretely, we discretize the current projection into a $128{\times}128$ grid, estimate a density histogram, and draw positions for the missing points by first sampling a grid cell according to those counts and then adding Gaussian jitter of half a cell width. 
This preserves the global density structure of the progressive projection while its size matches the ground-truth projection.

For the second challenge, instead of comparing absolute coordinates, we compare the \knn of each point in a progressive projection with ground-truth neighbor sets.
We define the per-point fidelity of point $i$ as $f_i = \frac{|N_k(i) \cap \mathcal{G}_i|}{k}$ where $N_k(i)$ is $i$'s \knn in the progressive projection and $\mathcal{G}_i$ the ground-truth neighbor set. 
The overall fidelity is $F = \frac{1}{M}\sum_{i=1}^{M'} f_i$.

To overcome the third challenge, we generate $R$ independent UMAP projections and take the union of their \knn sets as the ground-truth neighbor set for each point: $\mathcal{G}_i = \bigcup_{r=1}^{R} N_k^{(r)}(i)$, where $N_k^{(r)}(i)$ is the \knn set of point $i$ in the $r$-th UMAP run.
Empirically, we increase $r$ up to 300 UMAP runs and stop earlier when the average fraction of newly observed neighbors, normalized by $k$, falls below 0.2.
We defer the justification of this stopping rule to the supplementary material. 

In principle, the fidelity of a progressive projection can reach 1.0.
However, we observed that even a newly computed ground-truth projection achieves a fidelity score below 1.0 due to inherent stochasticity.
To account for this, we normalize the fidelity by the mean fidelity $F_{\text{UMAP}}$ of \ustd runs, yielding the \textbf{normalized fidelity}:
$\hat{F} = \frac{F}{F_{\text{UMAP}}}$.
A value of $\hat{F} = 1.0$ indicates that the progressive projection is as faithful as a newly computed ground-truth projection.

\subsection{Sanity Checks}

We first verify whether \ours produces final results consistent with \ustd when progressive scheduling is disabled.
Specifically, we run \ours using the Sequential strategy and compare its final recall and normalized fidelity with those obtained from \ustd (\cref{tab:sanity_check}).
\edit{Across most datasets, Sequential achieves high final recall and normalized fidelity, indicating that the modifications introduced by \ours largely preserve the final behavior of \ustd.
The two text embedding datasets show comparatively lower final recall, while still maintaining moderate to high normalized fidelity.}

\begin{figure}[t!]
    \centering
    \includegraphics[
        alt={},
        width=\linewidth
    ]{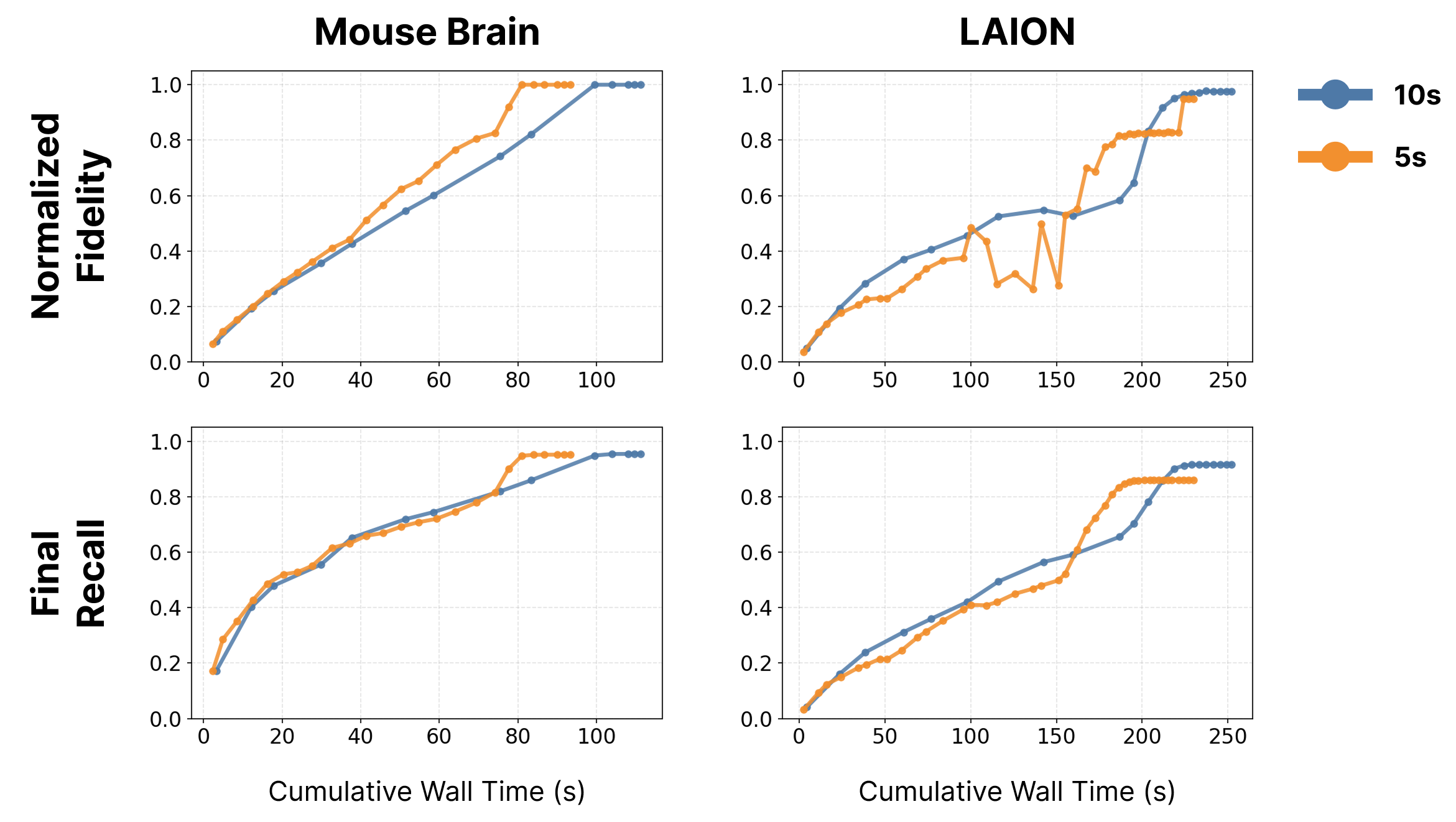}
    \caption{
        Comparison of two time-budget settings (5 seconds and 10 seconds) on Mouse Brain and LAION, showing normalized fidelity and final recall over cumulative wall time.}
    \vspace{-3mm}
    \label{fig:latency}
\end{figure}

\subsection{Progressive Behavior}
\textbf{Fidelity and Scheduling Strategies.}
The first and second rows of \cref{fig:strategies} show the normalized fidelity and final recall, respectively, of different scheduling strategies over cumulative wall time.
\seq is shown as a single point, as it does not produce intermediate \knn graphs or projections before completing the full computation.
In contrast, all progressive strategies expose intermediate results well before \seq completes.

\edit{Among the evaluated strategies, \updFoc shows the most balanced improvement in both normalized fidelity and final recall.
This behavior is consistent with the relative amount of work required by the three operation types in our pipeline: neighborhood refinement generally requires more \callUpd{} operations than the corresponding insertion or embedding operations.
Allocating additional budget to \callUpd{} therefore allows neighborhood refinement to better keep pace with newly inserted points.}

\edit{In contrast, \insFoc shows that normalized fidelity and final recall improve relatively slowly during insertion and then rise sharply once insertion is completed.
After this point, the budget previously allocated to insertion can be redirected to the remaining update and embedding operations, allowing the neighborhood graph and projection to rapidly catch up.
\embFoc shows a broadly similar pattern, although its trajectory generally lies between \insFoc and \updFoc.
Overall, these results suggest that scheduling mainly determines how well neighborhood refinement keeps pace with data insertion during progressive execution, while the differences among strategies become less pronounced after insertion is completed.}

\textbf{Time Budgets.}
We examine time budgets from two perspectives: (1) how faithfully the system adheres to a specified per-run budget, and (2) how different budget settings affect progressive behavior.
We use the \updFoc strategy for both analyses.

To assess the first aspect, we inspect the distribution of wall time between consecutive runs (third row of \cref{fig:strategies}).
Across datasets, the observed wall times remain close to the specified budget, although some runs terminate earlier when the number of available operations in the queues is insufficient.
This indicates that the runtime of \ours is predictable and can be controlled by adjusting the number of operations of each type, even with simple linear regression predictors.

\edit{For the second aspect, \cref{fig:latency} compares
$t_{\mathrm{budget}}=5$s and $10$s on Mouse Brain and LAION.
We initially expected the shorter budget to require a longer total execution time because more frequent runs could incur greater progressive overhead.
Contrary to this expectation, the 5-second setting completed earlier than the 10-second setting on both datasets.
However, halving the time budget does not necessarily double the number of runs, since the amount of available work varies throughout progressive execution and several runs terminate before exhausting the allocated budget.
Consequently, we did not observe a clear relationship between the per-run time budget and the total execution time.}

\edit{A more noticeable difference appears in normalized fidelity.
While the two budget settings exhibit relatively similar fidelity trajectories on Mouse Brain, the 5-second setting shows larger fluctuations on LAION.
This observation suggests that shorter time budgets may lead to less stable intermediate projection quality, particularly for larger-scale data.}

\begin{figure}[t!]
    \centering
    \includegraphics[width=\linewidth]{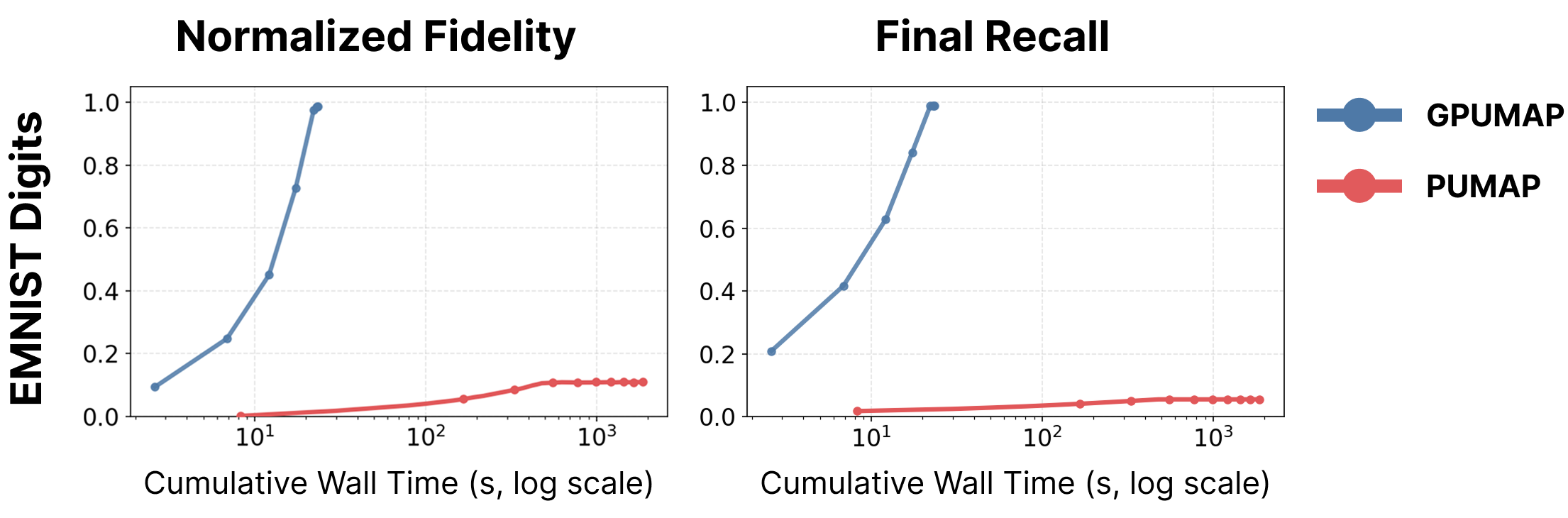}
    \caption{
    Comparison of GPUMAP and PUMAP on EMNIST Digits, showing normalized fidelity and final recall over cumulative wall time.}
    \label{fig:pumap}
    \vspace{-3mm}
\end{figure}

\begin{figure*}[!htb]
    \centering
    \includegraphics[
        alt={},
        width=\linewidth
    ]{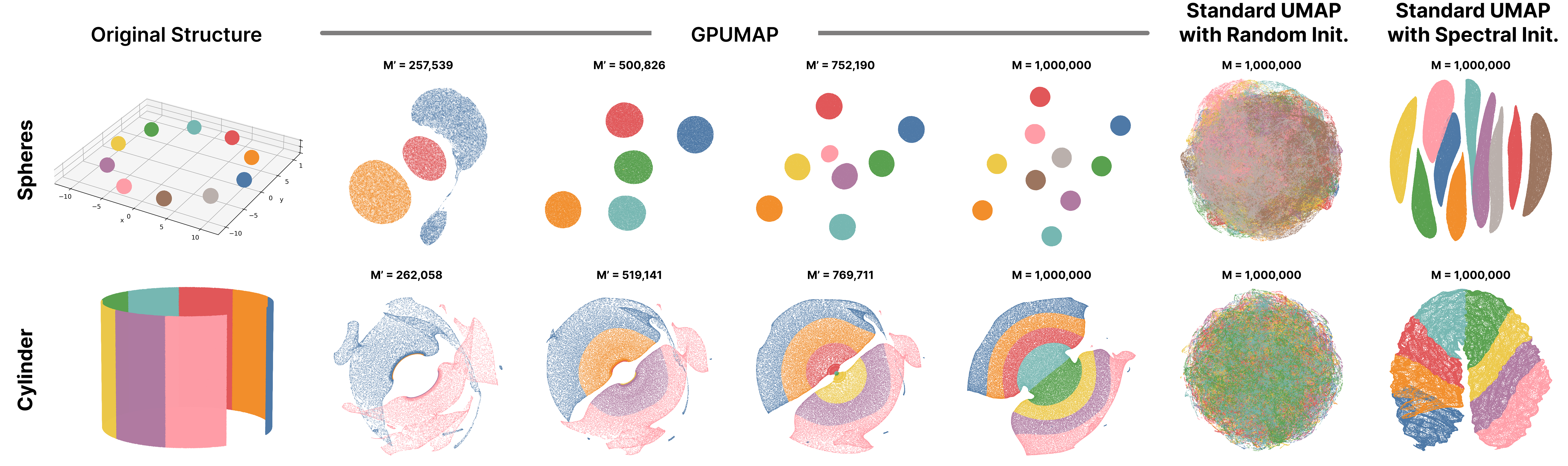}
    \caption{Progressive behavior of GPUMAP on synthetic datasets with controlled geometric structures and insertion orders. The left column shows the original 3D structures, the middle columns show intermediate GPUMAP projections as the number of inserted points \(M'\) increases, and the right columns show standard UMAP projections with random and spectral initialization for reference.}
    \vspace{-2mm}
    \label{fig:robustness}    
\end{figure*}

\textbf{Overhead.}
We next quantify the computational cost introduced by supporting progressive execution, using \seq as a non-progressive reference.
For \updFoc, the total runtime is 1.3$\times$, 1.6$\times$, and 1.8$\times$ that of \seq across the three datasets, respectively.
This overhead arises not only from coordination required for bounded-latency execution, such as CPU–GPU synchronization, but also because progressive execution performs additional insertion, update, and embedding operations to maintain reasonably high-fidelity intermediate results throughout the run.
In sum, progressive execution introduces a clear trade-off between throughput and the ability to access faithful intermediate results during execution.

\subsection{Comparison with Progressive UMAP}

\edit{We compare \ours with Progressive UMAP (PUMAP)~\cite{ko2020progressive}, a prior CPU-based progressive implementation of UMAP.
Because PUMAP becomes impractical for the larger datasets in our evaluation, we perform the comparison on EMNIST Digits, the smallest real-world dataset in our benchmark with 280K points.
PUMAP controls progressive computation using a fixed number of operations per batch rather than a wall-time budget.
Following its original two-stage execution scheme, which allocates a larger batch at initialization and smaller batches thereafter, we use an initial batch of 60,000 points followed by batches of 6,000 points to adapt the method to the larger EMNIST Digits dataset.
We otherwise follow the settings of the original implementation.}

\edit{As shown in \cref{fig:pumap}, \ours improves both normalized fidelity and final recall substantially faster than PUMAP.
\ours reaches near-reference levels for both metrics within tens of seconds, whereas PUMAP remains at considerably lower fidelity and recall even after substantially longer cumulative execution time.
Together with the difficulty of running the CPU-based implementation on our larger datasets, these results highlight the practical scalability advantage of \ours.}

\subsection{Robustness Assessment}

\cref{fig:robustness} illustrates how the projections change over time for the synthetic datasets under biased insertion orders.
\edit{Although inputs would typically be shuffled at random in practice, we deliberately construct extreme insertion orders to evaluate the robustness of progressive DR under unfavorable ordering conditions.}

\textbf{Spheres.}
\edit{Spheres consists of ten compact 3D spherical clusters, each containing 100,000 points, with their centers evenly arranged along a circle.
To introduce a strong temporal bias, we insert all points from each sphere contiguously before proceeding to the next sphere.
Under this ordering (\cref{fig:robustness}-Spheres),} we observe an incorrect split of a sphere in the early projection (e.g., the blue cluster in the first projection).
We attribute this artifact to stochastic initialization: when points from a new sphere (e.g., the red sphere) begin to arrive, they are randomly placed near the center, forming an emerging cluster.
Due to negative sampling, this newly formed structure exerts repulsive forces that can temporarily disrupt and split an existing cluster.
Despite this early instability, the split is progressively resolved in subsequent projections, and the embedding ultimately converges to a clean and well-separated layout across all ten spheres.

\textbf{Cylinder.}
\edit{Cylinder consists of points sampled from the lateral surface of a cylinder and labeled by angular sector, with one sector intentionally left empty to create a controlled geometric gap.
To create a strongly biased temporal ordering, we insert the sectors in an interleaved order $(1,9,2,8,\ldots)$, which separates spatially adjacent regions in the insertion sequence.
Under this ordering (\cref{fig:robustness}-Cylinder), the early projections already preserve the separation among the currently inserted sections.
As progression continues, the currently inserted sections remain separated until the intermediate regions that connect them are introduced; as these points arrive, the separated structures gradually become connected through the center, progressively revealing the global relationships among the data observed so far.
This suggests that the intermediate projections reflect the structure of the currently inserted subset reasonably well. 
At the same time, because newly inserted points are initialized near the center of the projection, points inserted earlier tend to be pushed outward as additional data arrive, suggesting that biased insertion continues to influence the progressive layout even while the projection continues to reflect the global relationships among the inserted regions.}

\del{In Cylinder (\cref{fig:robustness}-Cylinder), we observe a ribbon-like artifact in the early projection (e.g., the first projection).
Although spatially distant in 3D, the two sections (blue and pink) are projected close together, likely due to the curved manifold structure, a phenomenon not observed in Spheres.
As progression continues, newly inserted sections expand outward from the center, gradually revealing the circular organization of the data.
This behavior reflects the underlying geometry, where points inserted later are closer along the cylindrical surface.
The final projection exhibits artifacts such as fold-overs and interleaving of non-adjacent colored sections, indicating the impact of biased insertion.}

\textbf{Comparison with Standard UMAP.}
In \cref{fig:robustness}, we also show reference projections generated by \ustd.
Surprisingly, \ustd with random initialization fails to recover the underlying structure, reaffirming its sensitivity to initialization~\cite{kobak2021initialization}.
This behavior can be attributed to strong repulsive forces from negative sampling, which may dominate before meaningful local structures are established, leading to local minima that are difficult to escape.
In contrast, progressive DR maintains a manageable number of points at each iteration, leaving sufficient space in the projection for points to move and reorganize freely, resulting in more faithful projections.

% \todo{단순히 1분 기다려서 RAPIDS보는 걸로는 어려운, 10초가 가능해서 되는 시나리오 찾아서 적기 (전에 미팅노트에 3개 논의했음 잘 기억은 안남..): 1) transform 같은 데이터 정제 후 빠른 sanity check, 2) 하이퍼 파라미터 탐색, 3) 인터랙티브한 시스템에서? 맥락 복구 비용 이런걸로 집중을 유지하는 한계 내로 이득 주장? 4) 5M 짜리는 우리 장비에서 1분인데, cold start에서는 장비에 따라서 1분이 걸리는지 아닌지도 사용자가 모르고, 더 큰 데이터는 10분이 될 수 있다는 것?}
% \subsection{Example: Visualizing a 5M Dataset}
% \cref{fig:higgs} shows intermediate projections generated by \ours on the HIGGS dataset (5M points).
% We observe that \ours reveals a structure close to its final configuration at an early stage, within approximately one minute.
% The global arrangement visible at one minute (the second scatterplot in \cref{fig:higgs}) remains largely stable thereafter, suggesting that the method can provide early, meaningful results even at this scale.
% This demonstration also highlights limitations of the current implementation: \ours requires substantially more time than standard UMAP to process the full dataset.
% As discussed in \cref{sect:result}, this gap may stem from progressive overheads, such as data transfer and extra computations.

\section{Discussion}
\edit{Our evaluation highlights both the practical opportunities and behavioral characteristics of progressive DR.
In this section, we discuss how early intermediate projections can support interactive analysis, interpret projection fidelity over time, and summarize the limitations of the current approach.}

\subsection{Use Cases}

\edit{\textbf{Large-Scale Data Exploration.}
\cref{fig:higgs} demonstrates a use case of \ours on the HIGGS dataset with 5M points~\cite{higgs_280}.
While \ustd provides its first projection only after completing the full computation at 50.4 seconds, \ours produces intermediate projections as early as 8.42 seconds and continues to refine them throughout execution.
The purpose of these early results is not to replace the final projection, but to allow users to begin inspecting the data while the computation is still running.
By providing updates within a short and predictable latency, \ours can keep the analysis loop active rather than forcing users to wait for the full projection and resume their analysis afterward.
This distinction may become increasingly important when the completion time is not known in advance or grows with the dataset size.}

\edit{\textbf{Early Assessment of Data Quality and Transformations.}
DR is often applied after preprocessing steps such as feature transformation, normalization, or scaling, and errors or inappropriate choices in these steps can lead to undesirable projections.
With a conventional sequential workflow, such problems become visible only after the full computation has completed.
In contrast, an early projection from \ours can provide a rapid sanity check of the current input, allowing users to detect unexpected structures and revise the preprocessing pipeline before spending the full computation time on an unsuitable configuration.}

\edit{\textbf{Hyperparameter Search.}
UMAP projections can vary substantially with hyperparameter choices such as the number of neighbors, minimum distance, and distance metric.
When exploring such configurations, users often need an initial indication of how a parameter choice affects the emerging structure before deciding whether it is worth completing the computation.
Intermediate projections from \ours can therefore support a more iterative exploration process.}

\edit{\textbf{Tunable Update Frequency.}
In an integrated visualization pipeline, each update also incurs data transfer and rendering costs, which are not included in our experiments.
Our implementation lets users adjust the frequency of progressive updates to account for these costs.
An analysis may begin with a small time budget to obtain frequent early updates and establish an initial understanding of the data.
Once frequent inspection is no longer necessary, the budget can be increased to avoid the transfer and rendering costs of updates the user no longer needs.}

\subsection{Progressive Behavior over Time}
\begin{figure}[t!]
    \centering
    \includegraphics[width=\linewidth]{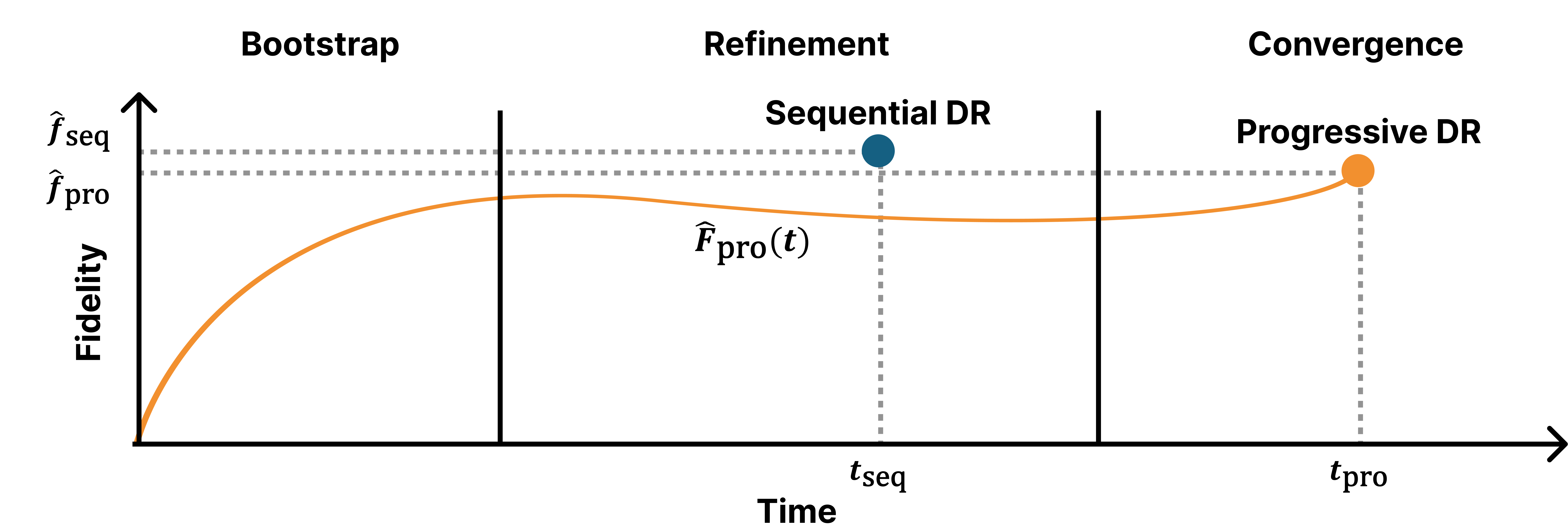}
    \caption{Projection fidelity of progressive DR over time.}
    \label{fig:fidelity_model}
    \vspace{-3mm}
\end{figure}

Based on our observation, we model the changes in the projection fidelity of progressive DR
(\cref{fig:fidelity_model}), in which the timeline is divided into three 
distinct phases: the \textit{Bootstrap Phase}\del{($< 1$ min)}, the 
\textit{Refinement Phase}, and the \textit{Convergence Phase}.

\textbf{Bootstrap Phase.}
During the \textit{Bootstrap Phase}, progressive DR rapidly constructs an initial partial projection as the first subsets of data are incorporated.
Although this early projection is still approximate, normalized fidelity generally increases rapidly during this stage, allowing useful structure to emerge before the full computation completes.
The duration of this phase varies across workloads rather than being
determined by a fixed amount of wall-clock time.

\textbf{Refinement Phase.}
During the \textit{Refinement Phase}, progressive DR continuously updates the projection by incrementally adding data points. The fidelity improvement during this phase is gradual, as the projection must balance consistency with previously placed points while accommodating newly introduced ones. We even observed temporary fidelity drops depending on the dataset and scheduling strategy.
Notably, sequential DR completes during this phase at time $t_{\text{seq}}$, at which point the fidelity gap $\Delta \hat{F}(t_{\text{seq}}) = f_{\text{seq}} -\hat{F}_{\text{pro}}(t_{\text{seq}})$ represents the disadvantage of progressive DR: had the user opted for sequential DR, they would have already obtained a complete projection.

\textbf{Convergence Phase.}
In the \textit{Convergence Phase}, the projection stabilizes as the majority of data points have been incorporated, rapidly approximating the sequential ground truth. The final fidelity difference $\Delta \hat{F}(t_{\text{pro}}) = f_{\text{seq}} - \hat{F}_{\text{pro}}(t_{\text{pro}})$ represents the loss attributable to incremental computation. 
Surprisingly, we did not observe a significant drop in final fidelity, which contrasts with prior work on progressive DR~\cite{jo2018panene}, where the $t$-SNE optimization became trapped in local minima and ultimately yielded lower fidelity than its sequential counterpart.
Instead, we found that the final fidelity of progressive DR is often higher than that of sequential DR.
A possible explanation is the difference in the initialization during projection: whereas sequential DR starts from random coordinates, progressive DR places newly inserted points based on the projection of previously processed points.
This more structured initialization provides a better starting point for optimization and helps the projection converge to a projection with high fidelity.

The ratio $\frac{t_{\text{pro}}}{t_{\text{seq}}}$ quantifies the computational overhead incurred by progressive execution. 
In our case, for the \updFoc strategy, we found this ratio to be approximately 1.6. 

Following this model, we summarize the performance characteristics of \ours below, which can serve as a baseline for future research:

\begin{itemize}
    % \item $\hat{F}_{\text{pro}}(t_{\text{boot}})$: \ours reaches a normalized fidelity above 0.3 within the first minute of execution, regardless of the dataset or scheduling strategy in our settings.

    \item $\Delta \hat{F}(t_{\text{seq}})$: on average, the normalized fidelity gap between sequential and progressive DR at the moment sequential DR completes is approximately 0.3.

    \item $\Delta \hat{F}(t_{\text{pro}})$: no significant fidelity loss is observed in the final projection due to progressive execution; in many cases, progressive DR achieved fidelity comparable to sequential DR.

    \item $\frac{t_{\text{pro}}}{t_{\text{seq}}}$: the total execution time of progressive DR is approximately 1.6 times that of sequential DR on average.
\end{itemize}

One implication of this temporal model is the potential for automatically adapting the time budget across different stages of progressive execution.
For example, a scheduler could favor frequent updates during the early stage, when rapid feedback is most valuable, and reduce the update frequency during periods of gradual refinement.
Such adaptation would require the system to infer the current stage from observable execution signals, since projection fidelity itself is not available during actual analysis.
Developing such phase-aware scheduling strategies remains an interesting direction for future work.

\subsection{Limitations}
Our approach also has several limitations.
First, we employ a simple linear regression model with a reserved margin to translate between time budgets and operation counts. While effective in practice, a more adaptive or learned cost model could better utilize the available budget.
Second, we use a fixed scheduling strategy with empirically determined weights. Although this choice facilitates comparisons across settings, it limits the scheduler's ability to adapt to dataset-specific characteristics or phase-dependent dynamics. 
Finally, our current implementation incurs additional overhead from generating intermediate projections between GPU operations, making it less efficient than a fully GPU-resident batch implementation. Reducing this communication overhead remains an important direction for future work.
\section{Conclusion}

We present \ours, a GPU-accelerated progressive implementation of UMAP that decomposes the original computation pipeline into three atomic operations, \callIns{}, \callUpd{}, and \callEmb{}, and organizes them into a controllable progressive execution model.
This design enables bounded-latency intermediate projections while scaling to million-point datasets on commodity GPU hardware.
Building on this implementation, we systematically investigate the behavior of progressive DR under diverse conditions, including different scheduling strategies, time budgets, and biased insertion orders.
Our results reveal trade-offs between runtime and projection fidelity, and establish an evaluation framework for progressive DR that we hope will inform
the design of future progressive visualization systems.

% \jj{모두 확인 완료. 좋습니다. 남아있는 것들만 수정합시다.}

%% if specified like this the section will be omitted in review mode
% \acknowledgments{%
% 	The authors wish to thank A, B, and C.
%   This work was supported in part by a grant from XYZ (\# 12345-67890).%
% }

\bibliographystyle{abbrv-doi-hyperref}

\bibliography{template}

\appendix % You can use the `hideappendix` class option to skip everything after \appendix
\crefalias{section}{appendix} % this is to make sure that cleverref switches to referring to Appx. X from here on

\section{About Appendices}
Refer to \cref{sec:appendices_inst} for instructions regarding appendices.

\section{Troubleshooting}
\label{appendix:troubleshooting}

\subsection{ifpdf error}

If you receive compilation errors along the lines of \texttt{Package ifpdf Error: Name clash, \textbackslash ifpdf is already defined} then please add a new line \verb|\let\ifpdf\relax| right after the \verb|\documentclass[journal]{vgtc}| call.
Note that your error is due to packages you use that define \verb|\ifpdf| which is obsolete (the result is that \verb|\ifpdf| is defined twice); these packages should be changed to use \verb|ifpdf| package instead.

\subsection{\texttt{pdfendlink} error}

Occasionally (for some \LaTeX\ distributions) this hyper-linked bib\TeX\ style may lead to \textbf{compilation errors} (\texttt{pdfendlink ended up in different nesting level ...}) if a reference entry is broken across two pages (due to a bug in \verb|hyperref|).
In this case, make sure you have the latest version of the \verb|hyperref| package (i.e.\ update your \LaTeX\ installation/packages).
If this package update does not fix the error for you, you can use the \verb|documentclass| option \verb|protectagainstpdfendlink| such as as follows:\\[1ex]
{\footnotesize\verb|\documentclass[journal]{vgtc,protectagainstpdfendlink}|}\\[1ex]
to set the \LaTeX/bib\TeX{} processing to avoid breaking reference entries in the list of references. Alternatively, you can revert back to \verb|\bibliographystyle{abbrv-doi}| (at the expense of removing hyperlinks from the bibliography) and then to try \verb|\bibliographystyle{abbrv-doi-hyperref}| again after some more editing.

\end{document}